\pdfoutput=1
\documentclass[runningheads]{llncs}
\usepackage{hyperref}
\usepackage{booktabs}
\usepackage{tabularx}
\usepackage{array}
\expandafter\def\csname ver@subfig.sty\endcsname{}
\usepackage{siunitx}
\usepackage{amsmath}
\usepackage{amssymb}
\usepackage{svg}
\usepackage{url}
\usepackage{nth}
\usepackage{mathtools}
\usepackage{multirow}
\usepackage{makecell}
\usepackage{microtype}
\usepackage[pages=some]{background}
\SetBgContents{\parbox{16cm}{
Accepted at the international conference on Information Processing in Medical Imaging (IPMI) 2019.\\
The final authenticated version is available online at \url{https://doi.org/10.1007/978-3-030-20351-1\_46
}.}}
\SetBgScale{1}
\SetBgAngle{0}
\SetBgOpacity{1}
\SetBgColor{black}
\SetBgPosition{current page.north west}
\SetBgHshift{10.4cm}
\SetBgVshift{-1.6cm}

\DeclareMathOperator*{\argmin}{arg\,min}
\newcolumntype{Y}{>{\centering\arraybackslash}X}
\newcolumntype{R}{>{\raggedleft\arraybackslash}X}
\newcolumntype{L}{>{\raggedright\arraybackslash}X}
\newcolumntype{n}{>{\hsize=0.95\hsize}Y}
\newcolumntype{m}{>{\hsize=1.2\hsize}Y}
\newcommand\notsotiny{\@setfontsize\notsotiny{6}{7}}

\newcommand{\bftab}{\fontseries{b}\selectfont}
\begin{document}
\title{Ultrasound Image Representation Learning by Modeling Sonographer Visual Attention}
\titlerunning{Ultrasound Image Representation Learning by Modeling Visual Attention}
\author{
Richard Droste\inst{1},
Yifan Cai\inst{1},
Harshita Sharma\inst{1},
Pierre Chatelain\inst{1},
\mbox{Lior Drukker\inst{2}},
Aris T.\ Papageorghiou\inst{2},
J.\ Alison Noble\inst{1}
}
\authorrunning{R.\ Droste et al.}
\institute{
Department of Engineering Science, \mbox{University of Oxford}, UK\\
\email{richard.droste@eng.ox.ac.uk}
\and
Nuffield Department of Women\textsc{\char13}s \& Reproductive Health, \mbox{University of Oxford, UK}
}
\maketitle
\BgThispage
\begin{abstract}
Image representations are commonly learned from class labels, which are a simplistic approximation of human image understanding.
In this paper we demonstrate that transferable representations of images can be learned without manual annotations by modeling human visual attention.
The basis of our analyses is a unique gaze tracking dataset of sonographers performing routine clinical fetal anomaly screenings.
Models of sonographer visual attention are learned by training a convolutional neural network (CNN) to predict gaze on ultrasound video frames through visual saliency prediction or gaze-point regression.
We evaluate the transferability of the learned representations to the task of ultrasound standard plane detection in two contexts.
Firstly, we perform transfer learning by fine-tuning the CNN with a limited number of labeled standard plane images.
We find that fine-tuning the saliency predictor is superior to training from random initialization, with an average F1-score improvement of 9.6\% overall and 15.3\% for the cardiac planes.
Secondly, we train a simple softmax regression on the feature activations of each CNN layer in order to evaluate the representations independently of transfer learning hyper-parameters.
We find that the attention models derive strong representations, approaching the precision of a fully-supervised baseline model for all but the last layer.
\end{abstract}
\section{Introduction}
When interpreting images, humans direct their attention towards semantically informative regions
\cite{Wu2014}.
This allocation of visual attention is typically quantified via the distribution of gaze points, which can be recorded with gaze tracking.
There has been great interest in developing models of human visual attention that, given an image, predict the likelihood that each pixel is fixated upon, hereafter referred to as \emph{visual saliency map}.
Currently, convolutional neural networks (CNNs) are the most effective visual attention models (VAMs) due to their ability to learn complex feature hierarchies through end-to-end training \cite{Borji2018}.
Here, we explore the following question:
\emph{To what extent can models of human visual attention transfer to tasks such as automatic image classification?}

We explore this question using the application of fetal anomaly ultrasound scanning.
The scan is performed during mid-pregnancy in order to detect fetal anomalies that require prenatal treatment and to determine the place, time and mode of birth.
Previous work related to this application has focused on detecting so-called ultrasound standard imaging planes through fully-supervised training of image classifiers \cite{Baumgartner2017,Cai2018c,Schlemper2018}.
Here, in contrast, we aim to learn transferable representations of the scan data without manual supervision by modeling sonographer visual attention.
To this end, we acquire the gaze of sonographers in real-time through unobtrusive gaze tracking alongside anomaly scan recordings.

Sonographer visual attention is modeled by training a CNN to predict gaze on random video frames.
We consider this to be \emph{self-supervised representation learning} since it does not require any manual annotations and gaze data is acquired fully automatically.
We extract high-resolution image features by introducing dilated convolutions \cite{Kruthiventi2015,Yu2016} into a
recently proposed image classification architecture \cite{Hu2017a}.
Two methods for training the model for gaze prediction are evaluated:
(i)
\emph{Visual saliency prediction:}
Ground truth visual saliency maps are generated and used as training targets \cite{Borji2018}.
(ii) 
\emph{Gaze-point regression:}
The approach of gaze-point regression \cite{Ngo2017} is much less explored in the literature but is simpler since it does not require explicit modeling of foveal vision for ground truth saliency map generation.
An existing mathematically differentiable method is based on a fully-connected layer \cite{Ngo2017} which does not scale well to high-resolution feature maps due to the exponentially increasing number of learnable parameters.
Here, we propose a method based on the soft-argmax algorithm by Levine et al.~\cite{Levine2015} with no additional learnable parameters compared to saliency prediction.

The learned representations are evaluated on the task of standard plane detection in two contexts.
(1) \emph{Transfer learning:}
We fine-tune the weights of the entire CNN with a limited number of training samples, thereby assessing the transferability of the learned representations in a realistic scenario.
\mbox{(2) \emph{Softmax regression:}}
We fix the weights of the CNN and train a simple softmax regression on the spatially average-pooled feature activations of each layer.
This procedure determines the generality of the representations independently of any transfer learning hyper-parameters.

\paragraph*{Related Work.}
Visual saliency predictors have previously been employed to aid computer vision tasks.
Cornia et al.\ \cite{Cornia2017a} use a pre-trained saliency predictor as an attention module within an image captioning architecture.
However, no representations are shared between the saliency predictor and the task-specific architecture.
Cai et al.\ \cite{Cai2018c} show that saliency prediction can aid fetal abdominal standard plane detection.
The authors fine-tune an existing standard plane detector \cite{Baumgartner2017} with manually labeled data, using saliency prediction as an auxiliary task and as an attention module.
In contrast, we show that transferable representations can be learned without manual annotations via visual attention modeling only.
Moreover, we evaluate our framework on full-length freehand clinical fetal anomaly scans instead of short sequences (sweeps) of the fetal abdomen.

Within the field of unsupervised representation learning, our method is most closely related to \emph{self-supervised learning}.
The general idea is to exploit ``free'' supervision signals, i.e., supervision signals that can be extracted from the data itself without any manual annotation, which is comparable to our approach of using automatically acquired gaze for supervision.
Specifically, representations are learned by either altering the data and inferring the alteration (e.g., spatial and color transformations \cite{Dosovitskiy2014}) or by predicting certain properties of the data that are withheld (e.g., the relative position of image patches \cite{Doersch2015} or the order of video frames \cite{Misra2016}).
However, all existing methods design artificial tasks that yield transferable representations as a ``by-product''.
Human gaze, in contrast, is inherently a strong prior for semantic information \cite{Wu2014}.

\paragraph*{Contributions.}
Our contributions are three-fold:
(1)
We propose an original framework for self-supervised image representation learning by modeling human visual attention.
The method does not require manual annotations, is generic, and has the potential to be applied in any setting where gaze tracking and image data can be acquired simultaneously.
To the best of our knowledge, this is the first attempt to study human visual attention modeling in the context of self-supervised representation learning;
(2)
we propose a method to regress gaze point coordinates via the soft-argmax algorithm, which is significantly simpler and more computationally efficient than the existing method by Ngo et al.~\cite{Ngo2017};
(3)
finally, we evaluate the attention models on the exemplary task of fetal anomaly ultrasound standard plane detection, both for transfer learning and as fixed feature extractors, thus demonstrating the applicability to a challenging real-world medical imaging task.
The framework is illustrated in \autoref{fig:method} a).

\section{Representation Learning by Modeling Visual Attention}
\label{sec:Method}
In this section we describe our method of learning image representations from video and gaze data in general terms.
\sloppy Let $\mathcal{X} \subset \mathbb{R}^{N_c\times H\times W}$ be the set of video frames with width $W$, height $H$ and $N_c$ channels and let \mbox{$\mathcal{P} = [0,W]{\times}[0,H]$} be the set of all valid gaze points.
Each frame $\mathbf{X} \in \mathcal{X}$ has a corresponding gaze point set \mbox{$G = \lbrace \mathbf{p}_i \,\vert\; \mathbf{p}_i \in \mathcal{P} \rbrace_{i=1}^{N_G}$} with $N_G\geq1$.
The dataset $\mathcal{D} = \left\lbrace \left( \mathbf{X}^{(t)}, G^{(t)}\right) \right\rbrace_{t=1}^{N_\mathtt{x}}$ consists of $N_\mathtt{x}$ pairs of video frames and gaze point sets.

\begin{figure}[tb]
{
\scriptsize
\sffamily
\begin{picture}(345,150)
\put(0,0){\includesvg[width=\textwidth]{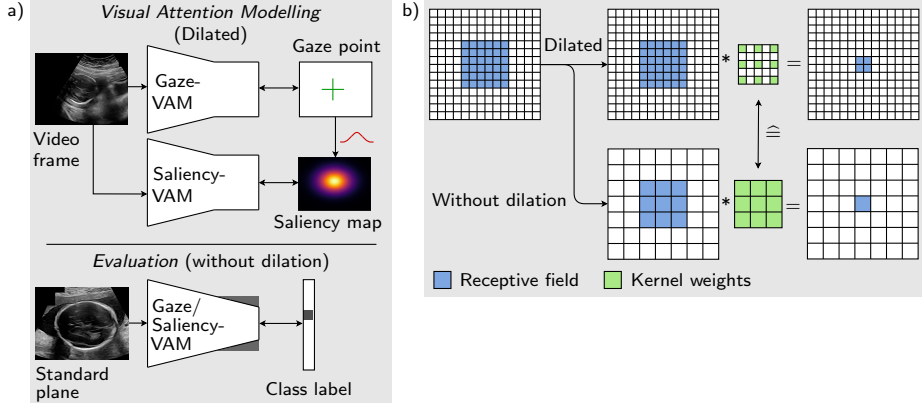}}
\end{picture}
}
\caption{
a) Illustration of our framework for learning and evaluating visual attention models (VAMs).
b) The upper part illustrates a dilated convolution after removed down-sampling operation.
When the down-sampling operation is reintroduced for classification as shown in the lower part, the dilation is removed from the kernel without changing the learned kernel weights.
The receptive field of the corresponding output neurons is unchanged and the operation is reversible.
}
\label{fig:method}
\label{fig:dilation}
\end{figure}

Let $\boldsymbol{f}_\theta: \mathcal{X} \rightarrow \mathbb{R}^{N_f\times\frac{1}{2^d}H \times \frac{1}{2^d} W}$ be a CNN with $N_f$ feature channels, \mbox{$2^d$-fold} spatial down-sampling and learnable parameters $\theta$.
The final, classification-specific operations (global pooling, fully-connected layers and softmax layers) are removed from the network at this stage.
In our experiments we use the SE-ResNeXt~\cite{Hu2017a} model, but any similar feed-forward CNN is suitable.
Since such models are designed for image classification, they perform strong down-sampling in order to increase the receptive field of the higher-level neurons and to reduce computational complexity.
In contrast, for visual attention modeling, it is desirable to preserve spatial information throughout the network.
Consequently, we remove the last $N_D$ down-sampling operations, i.e., max-pooling or strided convolutions.
However, this modification reduces the receptive field of the subsequent neurons.
If the down-sampling operations were reintroduced to restore the original architecture and use the representations for classification tasks, the learned weights would be invalid.
Therefore, the 3$\times$3 convolutions after the removed down-sampling operations are replaced with \emph{dilated} 3$\times$3 convolutions \cite{Yu2016} such that each convolutional kernel maintains the same receptive field as in the original architecture, as illustrated in \autoref{fig:dilation} b).
Formally, given a matrix $\mathbf{M}$ and the kernel \mbox{$k:[-r,r]^2 \cap \mathbb{Z}^2 \rightarrow \mathbb{R}$ of size $(2r + 1)^2$}, the $l$-fold dilated convolution operator $\ast_l$ is defined as:
\begin{equation}
(\mathbf{M}\ast_l k)_{i,j} = \sum_{n=\text{-}r}^r \sum_{m=\text{-}r}^r \mathbf{M}_{i+ln,j+lm} \;k(n,m)
\end{equation}
The resulting dilated CNN $\boldsymbol{f}^\oplus_\theta$ has the increased output resolution of $(H_D, W_D) \coloneqq (\frac{1}{2^{d-N_D}} H, \frac{1}{2^{d-N_D}} W)$.
Next, we want to reduce the high-dimensional feature activations to a single probability map that can be used to model visual attention.
Hence, a series of \emph{adaptation layers} consisting of a 7$\times$7 depthwise convolution and several 1x1 convolutions is appended that outputs a single activation map $\mathbf{A} \in \mathbb{R}^{H_D\times W_D}$.
A probability map $\mathbf{\hat{S}}$ is then computed by applying a \emph{softmax} across the activations with $\hat{S}_{i,j} = e^{A_{i,j}}/ \sum_{i,j}e^{A_{i,j}}$.

We investigate two methods of training the CNN to predict gaze in a differentiable, and therefore end-to-end trainable, manner: \emph{visual saliency prediction} and \emph{gaze-point regression}.

\paragraph*{Visual Saliency Prediction.}
\label{sec:Method::Saliency}
Given an image and a gaze point set $(\mathbf{X}, G) \in \mathcal{D}$, the idea is to generate a \emph{visual saliency map} $\mathbf{S}\in\;]0,1]^{H\times W}$, where $S_{i,j}$ is the probability that pixel $X_{i,j}$ is fixated upon.
The saliency map is then used as the target for the predicted probability map $\mathbf{\hat{S}}$.
We generate $\mathbf{S}$ as a sum of Gaussians around the gaze points in $G$, normalized such that $\sum_{i,j}S_{i,j} = 1$.
The standard deviation of the Gaussians is equivalent to ca.\ \SI{1}{\degree} visual angle to account for the radius of visual acuity and the uncertainty of the eye tracker measurements \cite{Chatelain2018}.
Next, the saliency map is downscaled to the size of $\mathbf{\hat{S}}$, yielding the training target $\mathbf{S}^\ast \in \;]0,1]^{H_D\times W_D}$.
Finally, the training loss is computed via the Kullback-Leibler divergence (KLD) between the predicted and the downscaled true distribution:
\begin{equation}
\mathcal{L}_s(\mathbf{S}^\ast, \hat{\mathbf{S}}) = D_\text{KL}(\mathbf{S}^\ast \|\, \hat{\mathbf{S}}) = \sum_{i,j} S^\ast_{i,j} \cdot (log(S^\ast_{i,j}) - log(\hat{S}_{i,j}))
\end{equation}

\paragraph*{Gaze-Point Regression.}
\label{sec:Method::Gaze}
We propose a method for reducing $\mathbf{\hat{S}}$ to a single gaze point in order to compare it to the true gaze points.
This eliminates the need to model the probability distribution of gaze points via a visual saliency map.
First, $\mathbf{\hat{S}}$ is transformed into image coordinates via the soft-argmax algorithm \cite{Levine2015}.
With $\boldsymbol{g}(i,j) \coloneqq \big(\frac{j-0.5}{W_D}\,W,\,\frac{i-0.5}{H_D}\,H\big)$ as the function that maps entry $(i, j)$ of $\mathbf{\hat{S}}$ to its corresponding point on the image plane, the predicted gaze point $\mathbf{\hat{p}}$ is computed as the expected value of the probability mass function defined by $\mathbf{\hat{S}}$:
\begin{equation}
\mathbf{\hat{p}} = \sum_{i,j} \hat{S}_{i,j}\: \boldsymbol{g}(i, j)
\end{equation}
Next, the target gaze point $\mathbf{p}^\ast$ is obtained from the gaze point set $G$ via the geometric median:
\begin{equation}
\mathbf{p}^\ast = \argmin_{\mathbf{p}^\ast\in[0,W]\times[0,H]} \sum_{p_i\in G} \left\| \mathbf{p}_i - \mathbf{p}^\ast \right\|_2
\end{equation}
This reduction is justified by the fact that the gaze points on each frame tend to be highly localized due to the short frame period (ca.\ \SI{33}{\milli\s}).
Finally, the training loss is obtained as $\mathcal{L}_g(\mathbf{p}^\ast, \mathbf{\hat{p}}) = \left\| \mathbf{p}^\ast - \mathbf{\hat{p}} \right\|_2$, i.e., the Euclidean distance between the predicted and the target gaze point.

\begin{table}[t!]
\caption{
SE-ResNeXt-50 (half-width) \cite{Xie2016} and SonoNet-64 \cite{Baumgartner2017} (variant of VGG-16 \cite{Simonyan2014}) architectures.
Convolutional layers are denoted as `conv \textless kernel-size\textgreater, \textless output-channels\textgreater[, \emph{\textless $C=$cardinality\textgreater}]', where cardinality is the number of grouped convolutions.
SE modules are denoted as `$fc$' followed by the dimensions of the corresponding fully-connected layers.
Scales in parentheses correspond to the dilated networks for attention modeling.
The lower part of the table shows the heads for attention modeling and classification, respectively.
}
\label{tab:models}
\scriptsize
\centering
\renewcommand*{\arraystretch}{1.1}
\begin{tabularx}{\textwidth}{p{18mm} | p{15mm}  L |  p{15mm}  p{28mm}}
\noalign{\hrule height 1pt}
& \multicolumn{2}{c|}{{\bftab SE-ResNeXt-50} (half-width, 7.4M parameters)}& \multicolumn{2}{c}{{\bftab SonoNet-64} (14.9M parameters)}\\
\cline{2-5}
Layer name& \multicolumn{1}{l}{Scale}& \multicolumn{1}{l|}{Layers}& \multicolumn{1}{l}{Scale}&\multicolumn{1}{l}{Layers}\\
\noalign{\hrule height 1pt}
\multirow{2}{*}{layer 1}& $224\times288$ & conv, $7\times 7$, 64, stride 2&
\multirow{2}{*}{$224\times288$}&
\multirow{2}{*}{$[$conv, $3\times3$, \leavevmode\phantom{0}64$]\times 2$}\\
&$112\times144$& max pool, $3\time 3$, stride 2&&\\
\noalign{\hrule height 0.5pt}
layer 2& $56\times72$&
$\left[\begin{array}{l}
    \text{conv, $1\times 1$, \leavevmode\phantom{0}64} \\
    \text{conv, $3\times 3$, \leavevmode\phantom{0}64, $C=16$} \\
    \text{conv, $1\times 1$, 128} \\
    \text{\emph{fc}, $[8, 128]$} \\
\end{array}\right]\times 3$
&$112\times144$& $[$conv, $3\times3$, 128$]\times 2$\\
\noalign{\hrule height 0.5pt}
layer 3& $28\times 36$&
$\left[\begin{array}{l}
    \text{conv, $1\times 1$, 128} \\
    \text{conv, $3\times 3$, 128, $C=16$} \\
    \text{conv, $1\times 1$, 256} \\
    \text{\emph{fc}, $[16, 256]$} \\
\end{array}\right]\times 4$
& $56\times72$&  $[$conv, $3\times3$, 256$]\times 3$\\
\noalign{\hrule height 0.5pt}
layer 4& \makecell[l]{$14\times 18$\\($28\times 36$)}&
$\left[\begin{array}{l}
    \text{conv, $1\times 1$, 256} \\
    \text{conv, $3\times 3$, 256, $C=16$} \\
    \text{conv, $1\times 1$, 512} \\
    \text{\emph{fc}, $[32, 512]$} \\
\end{array}\right]\times 6$
& $28\times36$&  $[$conv, $3\times3$, 512$]\times 3$\\
\noalign{\hrule height 0.5pt}
layer 5& \makecell[l]{$7\times 9$\\($28\times 36$)}&
$\left[\begin{array}{l}
    \text{conv, $1\times 1$, 512} \\
    \text{conv, $3\times 3$, 512, $C=16$} \\
    \text{conv, $1\times 1$, 1024} \\
    \text{\emph{fc}, $[64, 1024]$} \\
\end{array}\right]\times 3$
& $14\times18$&  $[$conv, $3\times3$, 512$]\times 3$\\
\noalign{\hrule height 0.5pt}\\[-8pt]
\noalign{\hrule height 0.5pt}
\makecell[l]{adaptation\\(attention)}
& $28\times 36$& \makecell[l]{conv $7\times 7$, 1024, $C=1024$\\$[$conv $1\times 1$, 256$]\times 2$\\conv $1\times 1$, 1}& \quad--- & --- \\
\noalign{\hrule height 0.5pt}\\[-8pt]
\noalign{\hrule height 0.5pt}
\makecell[l]{adaptation\\(classification)}
& $7\times 9$& \makecell[l]{conv $1\times 1$, 256\\conv $1\times 1$, $N_C$\\avg.\ pool and softmax}& $14\times 18$& \makecell[l]{conv $1\times 1$, 256\\conv $1\times 1$, $N_C$\\avg.\ pool and softmax}\\
\noalign{\hrule height 1pt}
\end{tabularx}
\end{table}

\section{Experiments}
\label{sec:Experiments}
\paragraph*{Data.}
We acquired a novel dataset of clinical fetal ultrasound exams with real-time sonographer gaze tracking data.
The exams are performed on a GE Voluson E8 scanner (General Electric, USA) while the video signal of the machine's monitor is recorded lossless at \SI{30}{\Hz}.
Gaze is simultaneously recorded at \SI{90}{\Hz} with a Tobii Eye Tracker 4C (Tobii, Sweden).
Ethics approval was obtained for data recording and data are stored according to local data governance rules.
For our experiments, we use 135 fetal anomaly scans, which are randomly split into three equally sized subsets for cross-validation.

\paragraph*{CNN Architecture.}
Recent empirical evidence suggests that ImageNet performance is strongly correlated with performance on other vision tasks \cite{Kornblith2018}.
Therefore, we base our CNN on SE-ResNeXt \cite{Hu2017a}, a ResNet-style model with aggregated convolutions and channel recalibration (\emph{squeeze-and-excitation}, short \emph{SE}) modules, which won the 2017 ImageNet classification competition.
For attention modeling, layers 4 and 5 are dilated as described in \autoref{sec:Method}.
In preliminary experiments we found that halving the number of feature channels (except for layer 0) greatly reduced the computational cost without performance losses on our dataset.
The resulting architecture is summarized in column 1 of \autoref{tab:models}.
Column 2 shows SonoNet-64 \cite{Baumgartner2017}, which we use as a reference for standard plane detection since the authors published network weights trained on over 22k standard plane images.

\subsection{Visual Attention Modeling}
\subsubsection*{Experimental methods.}
Two visual attention models (VAMs) were trained on the ultrasound video and gaze data as described in \autoref{sec:Method}, namely a visual saliency predictor (\emph{Saliency-VAM}) and a gaze-point regressor (\emph{Gaze-VAM}).
For pre-processing, all video frames that did not correspond to 2D B-mode live scanning (e.g., Doppler, 3D/4D or frozen frames) or without gaze data were discarded.
Further, all but every \nth{8} frame were discarded to reduce temporal redundancy, resulting in a total of \SI{403070} video frames.
Next, the frames were cropped down to the region of the actual ultrasound image.
Data augmentation was performed by uniformly sampling sub-crops of 70-90\% side length that contained the gaze points, random horizontal flipping, and varying gamma and brightness by $\pm25$\%.
Finally, the frames were down-sampled to a size of $224{\times}288$ pixels and normalized to zero-mean and unit-variance.

Both attention models were trained via stochastic gradient descent (SGD) with momentum of 0.9, weight decay of $10^{-4}$ and mini-batch size of 32.
The Saliency-VAM was trained for 8 epochs at a learning rate (LR) of 0.1 while the Gaze-VAM converged more slowly and was trained for 10 epochs at a LR of 0.01.
In each case, the LR was decayed by a factor of 10 for the final two epochs.
All experiments were implemented in the PyTorch framework.
Each training run was performed in \SIrange{9}{16}{\hour} on a single Nvidia GTX 1080 Ti.

\begin{table}[tb]
\caption{
Results of visual saliency prediction and gaze-point regression compared to static baselines (mean $\pm$ standard deviation).
Next to the training loss (KLD), the Saliency-VAM is evaluated on the metrics normalized scanpath saliency (NSS), AUC-Judd, Pearson's correlation coefficient (CC) and histogram intersection (SIM) (for references see \cite{Borji2018}).
Best values are marked bold.
}
\label{tab:gaze_scores}
\centering
\begin{small}
\renewcommand*{\arraystretch}{1.1}
\begin{tabularx}{\textwidth}{l | *{5}{Y} | Y}
\noalign{\hrule height 1pt}
&\multicolumn{5}{c|}{Saliency-VAM}& \multicolumn{1}{c}{Gaze-VAM}\\
\cline{2-7}
& \multicolumn{1}{c}{KLD}& \multicolumn{1}{c}{NSS}& \multicolumn{1}{c}{AUC [\%]}& \multicolumn{1}{c}{CC [\%]}& \multicolumn{1}{c|}{SIM [\%]}& \multicolumn{1}{c}{$\ell_2$-norm}\\
\noalign{\hrule height 0.5pt}
Static& 3.41 {\scriptsize $\pm$0.02}& 1.39 {\scriptsize $\pm$0.05}& 85.9 {\scriptsize $\pm$0.3}& 14.9 {\scriptsize $\pm$0.4}& \leavevmode\phantom{0}8.5 {\scriptsize $\pm$0.1}& 54.4 {\scriptsize $\pm$0.6}\\
Learned& {\bftab 2.43} {\scriptsize $\pm$0.03}& {\bftab 4.03} {\scriptsize $\pm$0.05}& {\bftab 96.7} {\scriptsize $\pm$0.2}& {\bftab 31.6} {\scriptsize $\pm$0.3}& {\bftab 18.5} {\scriptsize $\pm$0.2}& {\bftab 27.4} {\scriptsize $\pm$0.4}\\
\noalign{\hrule height 1pt}
\end{tabularx}
\end{small}
\end{table}

\begin{figure}[tb]
{
\centering
\scriptsize
\sffamily
\includesvg[width=0.88\textwidth]{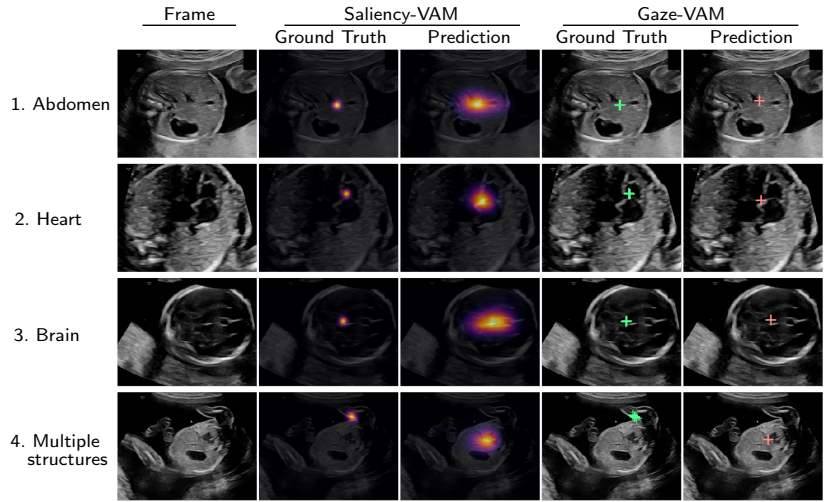}
}
\caption{
Visual saliency and gaze point predictions with corresponding ground truths for representative validation set frames.
}
\label{fig:sal_gaze_examples}
\end{figure}

\subsubsection*{Results.}
\autoref{tab:gaze_scores} summarizes the quantitative evaluation of the attention models.
The static baseline for the Saliency-VAM is the normalized sum of all ground truth saliency maps.
The baseline for the Gaze-VAM is the geometric median of all gaze points.
The learned models clearly outperform the static baselines on every metric.
\autoref{fig:sal_gaze_examples} shows visual saliency and gaze point predictions for four representative frames from the validation set.
Frames 1.-3.\ each contain one anatomical structure and show examples of accurate prediction.
Frame 4.\ contains several structures, which creates ambiguity.

\subsection{Fetal Anomaly Standard Plane Detection}
\subsubsection*{Experimental methods.}
For comparison with Baumgartner et al.~\cite{Baumgartner2017}, we consider the same 13 standard plane classes and ``background'' class, except that our data contains the three vessels and trachea view (3VT) which is similar to their three-vessel view (3VV).
From the available 135 anomaly scans, we obtained a total of 1129 standard plane frames with 62 to 148 samples per class (a plane may be acquired twice or may be skipped in a scan).
Moreover, we sampled 1127 background frames in the vicinity of the standard planes.
The same scan-level three-fold cross-validation split as for attention modeling was applied.
For pre-processing, frames were cropped to the ultrasound image region as for attention modeling.
The images were then augmented by random horizontal flipping, rotation by $\pm$\SI{10}{\degree}, varying the aspect-ratio by $\pm$10\%, sampling a sub-crop of 95-100\% side length,
and varying gamma and brightness by $\pm25$\%.
As before, the images were down-sampled and normalized.

The trained visual attention models were fine-tuned (FT) on the standard plane detection task, yielding \emph{Saliency-FT} and \emph{Gaze-FT}.
Moreover, two baselines were generated: A SE-ResNeXt model trained from random initialization and a fine-tuned SonoNet-64 (\emph{SonoNet-FT}).
Each epoch consisted of 1024 randomly sampled images.
Analogous to Baumgartner~et~al., we overcome the class imbalance problem by sampling images from each standard plane class with the same frequency and sampling one background image per standard plane image.
Fine-tuning was performed via SGD with momentum of 0.9, weight decay of $5{\times}10^{-4}$, mini-batch size of 16 and a cross-entropy loss.
The attention models were fine-tuned for 50 epochs with a LR of 0.01, decayed by a factor of 10 at epochs 20 and 35.
For the randomly initialized model, the LR was increased by a factor of 4.
The SonoNet model was initialized with pre-trained weights published by the authors and fine-tuned for 25 epochs with a LR of 0.01, decayed at epochs 10 and 20.
Longer training or higher learning rates led to overfitting for the latter two models due to the relatively small number of training samples.
Due to the class imbalance, the overall precision, recall and F1-scores were computed as \emph{macro-averages}, i.e., the average of the scores per standard plane.

Besides fine-tuning, we trained a multinomial logistic regression (softmax regression) on the spatially average-pooled feature activations of each layer of the attention models and two baselines: an SE-ResNeXt model with random weights and the pre-trained SonoNet model.
For each regression, the entire respective training set was sampled without augmentation.
The classifier was trained with the L-BFGS solver and balanced class weights.
The L2 regularization parameter was selected for each regression from a range of 16 logarithmically spaced values from $10^{-5}$ to $10^1$ based on the validation F1-score.

\begin{table}[tb]
\caption{
Standard plane detection results after fine-tuning (mean $\pm$ standard deviation [\%]).
\emph{Rand.\ Init.\ }denotes the SE-ResNeXt model trained from scratch.
The best score among the first three models is marked in bold.
Scores of the fine-tuned SonoNet that exceed all three models are marked in bold as well.
The literature SonoNet scores are given in parenthesis.
}
\label{tab:scores}
\centering
\renewcommand*{\arraystretch}{1}
\begin{tabularx}{\textwidth}{L | *{3}{n} | n | m}
\noalign{\hrule height 1pt}
& Rand.\ Init.&  Gaze-FT& Saliency-FT& $\Delta$(Saliency, Rand.\ Init.)& SonoNet-FT (Lit.\ value \cite{Baumgartner2017})\\
\noalign{\hrule height 0.5pt}
Precision& 70.4 {\scriptsize $\pm$\leavevmode\phantom{0}2.3}& 67.2 {\scriptsize $\pm$\leavevmode\phantom{0}3.4}& {\bftab 79.5} {\scriptsize $\pm$\leavevmode\phantom{0}1.7}& \leavevmode\phantom{0}9.1 {\scriptsize $\pm$\leavevmode\phantom{0}2.1}& {\bftab 82.3} {\scriptsize $\pm$\leavevmode\phantom{0}1.3} (81)\\
Recall& 64.9 {\scriptsize $\pm$\leavevmode\phantom{0}1.6}& 57.3 {\scriptsize $\pm$\leavevmode\phantom{0}4.5}& {\bftab 75.1} {\scriptsize $\pm$\leavevmode\phantom{0}3.4}& 10.2 {\scriptsize $\pm$\leavevmode\phantom{0}1.9}& {\bftab 87.3} {\scriptsize $\pm$\leavevmode\phantom{0}1.1} (86)\\
F1-score& 67.0 {\scriptsize $\pm$\leavevmode\phantom{0}1.3}& 60.7 {\scriptsize $\pm$\leavevmode\phantom{0}3.9}& {\bftab 76.6} {\scriptsize $\pm$\leavevmode\phantom{0}2.6}& \leavevmode\phantom{0}9.6 {\scriptsize $\pm$\leavevmode\phantom{0}2.1}& {\bftab 84.5} {\scriptsize $\pm$\leavevmode\phantom{0}0.9} (83)\\
\noalign{\hrule height 0.5pt}
\multicolumn{4}{l}{\emph{F1-scores per class:}}& \multicolumn{1}{c}{$\downarrow$}\\
RVOT& 37.9 {\scriptsize $\pm$\leavevmode\phantom{0}3.8}& 30.4 {\scriptsize $\pm$\leavevmode\phantom{0}4.9}& {\bftab 58.7} {\scriptsize $\pm$\leavevmode\phantom{0}2.7}& 20.8 {\scriptsize $\pm$\leavevmode\phantom{0}5.5}& {\bftab 71.2} {\scriptsize $\pm$\leavevmode\phantom{0}2.8}\\
LVOT& 30.3 {\scriptsize $\pm$\leavevmode\phantom{0}4.7}& 25.8 {\scriptsize $\pm$\leavevmode\phantom{0}5.9}& {\bftab 48.6} {\scriptsize $\pm$\leavevmode\phantom{0}3.3}& 18.4 {\scriptsize $\pm$\leavevmode\phantom{0}7.3}& {\bftab 69.9} {\scriptsize $\pm$\leavevmode\phantom{0}5.3}\\
4CH& 43.1 {\scriptsize $\pm$\leavevmode\phantom{0}6.7}& 33.5 {\scriptsize $\pm$\leavevmode\phantom{0}8.5}& {\bftab 57.3} {\scriptsize $\pm$10.8}& 14.2 {\scriptsize $\pm$11.9}& {\bftab 75.7} {\scriptsize $\pm$\leavevmode\phantom{0}9.1}\\
Kidneys& 71.4 {\scriptsize $\pm$\leavevmode\phantom{0}5.5}& 68.5 {\scriptsize $\pm$12.1}& {\bftab 84.7} {\scriptsize $\pm$\leavevmode\phantom{0}6.3}& 13.3 {\scriptsize $\pm$\leavevmode\phantom{0}5.7}& 81.0 {\scriptsize $\pm$\leavevmode\phantom{0}5.0}\\
Profile& 77.5 {\scriptsize $\pm$\leavevmode\phantom{0}7.2}& 61.7 {\scriptsize $\pm$\leavevmode\phantom{0}7.7}& {\bftab 87.2} {\scriptsize $\pm$\leavevmode\phantom{0}7.5}& \leavevmode\phantom{0}9.7 {\scriptsize $\pm$\leavevmode\phantom{0}3.7}& {\bftab 88.1} {\scriptsize $\pm$\leavevmode\phantom{0}4.5}\\
Lips& 76.7 {\scriptsize $\pm$\leavevmode\phantom{0}2.6}& 74.2 {\scriptsize $\pm$\leavevmode\phantom{0}7.3}& {\bftab 85.6} {\scriptsize $\pm$\leavevmode\phantom{0}4.5}& \leavevmode\phantom{0}8.8 {\scriptsize $\pm$\leavevmode\phantom{0}6.7}& {\bftab 92.9} {\scriptsize $\pm$\leavevmode\phantom{0}0.8}\\
Brain (Cb.)& 84.9 {\scriptsize $\pm$\leavevmode\phantom{0}7.0}& 83.2 {\scriptsize $\pm$\leavevmode\phantom{0}1.5}& {\bftab 93.7} {\scriptsize $\pm$\leavevmode\phantom{0}4.6}& \leavevmode\phantom{0}8.8 {\scriptsize $\pm$\leavevmode\phantom{0}2.3}& 92.8 {\scriptsize $\pm$\leavevmode\phantom{0}1.1}\\
3VT& 50.4 {\scriptsize $\pm$\leavevmode\phantom{0}1.9}& 47.1 {\scriptsize $\pm$\leavevmode\phantom{0}7.1}& {\bftab 58.3} {\scriptsize $\pm$\leavevmode\phantom{0}7.1}& \leavevmode\phantom{0}7.9 {\scriptsize $\pm$\leavevmode\phantom{0}5.6}& {\bftab 77.9} {\scriptsize $\pm$\leavevmode\phantom{0}1.6}\\
Brain (Tv.)& 86.1 {\scriptsize $\pm$\leavevmode\phantom{0}7.7}& 88.8 {\scriptsize $\pm$\leavevmode\phantom{0}2.8}& {\bftab 92.9} {\scriptsize $\pm$\leavevmode\phantom{0}5.0}& \leavevmode\phantom{0}6.8 {\scriptsize $\pm$\leavevmode\phantom{0}2.8}& 92.1 {\scriptsize $\pm$\leavevmode\phantom{0}4.5}\\
Spine (cor.)& 72.9 {\scriptsize $\pm$\leavevmode\phantom{0}3.6}& 57.2 {\scriptsize $\pm$\leavevmode\phantom{0}2.8}& {\bftab 79.0} {\scriptsize $\pm$\leavevmode\phantom{0}3.7}& \leavevmode\phantom{0}6.1 {\scriptsize $\pm$\leavevmode\phantom{0}6.2}& {\bftab 90.3} {\scriptsize $\pm$\leavevmode\phantom{0}4.9}\\
Abdominal& 67.9 {\scriptsize $\pm$\leavevmode\phantom{0}5.1}& 60.8 {\scriptsize $\pm$\leavevmode\phantom{0}6.7}& {\bftab 72.9} {\scriptsize $\pm$\leavevmode\phantom{0}2.9}& \leavevmode\phantom{0}5.0 {\scriptsize $\pm$\leavevmode\phantom{0}3.7}& {\bftab 85.0} {\scriptsize $\pm$\leavevmode\phantom{0}1.4}\\
Spine (sag.)& 86.5 {\scriptsize $\pm$\leavevmode\phantom{0}3.5}& 80.2 {\scriptsize $\pm$\leavevmode\phantom{0}2.5}& {\bftab 89.1} {\scriptsize $\pm$\leavevmode\phantom{0}2.1}& \leavevmode\phantom{0}2.7 {\scriptsize $\pm$\leavevmode\phantom{0}5.2}& {\bftab 91.6} {\scriptsize $\pm$\leavevmode\phantom{0}2.5}\\
Femur& 85.7 {\scriptsize $\pm$\leavevmode\phantom{0}2.0}& 77.7 {\scriptsize $\pm$\leavevmode\phantom{0}0.1}& {\bftab 87.6} {\scriptsize $\pm$\leavevmode\phantom{0}1.3}& \leavevmode\phantom{0}1.9 {\scriptsize $\pm$\leavevmode\phantom{0}1.5}& {\bftab 89.5} {\scriptsize $\pm$\leavevmode\phantom{0}1.8}\\
Background& 85.2 {\scriptsize $\pm$\leavevmode\phantom{0}0.9}& 83.3 {\scriptsize $\pm$\leavevmode\phantom{0}0.7}& {\bftab 89.0} {\scriptsize $\pm$\leavevmode\phantom{0}0.4}& \leavevmode\phantom{0}3.8 {\scriptsize $\pm$\leavevmode\phantom{0}1.2}& {\bftab 90.3} {\scriptsize $\pm$\leavevmode\phantom{0}0.4}\\
\noalign{\hrule height 1pt}
\multicolumn{6}{l}{
\makecell[{{p{\textwidth}}}]{
\scriptsize
RVOT: right ventricular outflow tract; LVOT: left ventricular outflow tract; 4CH: four chamber view; 3VT: three vessel and trachea view; Brain (Cb.): brain cerebellum suboccipitobregmatic plane; Brain (Tv.): brain transventricular plane; Cor.: Coronal plane; Sag.: Sagittal plane.
}}\\
\noalign{\hrule height 1pt}
\end{tabularx}
\end{table}

\begin{figure}[t]
{
\begin{picture}(347,108)
\put(0,0){\includesvg[width=1.008\textwidth]{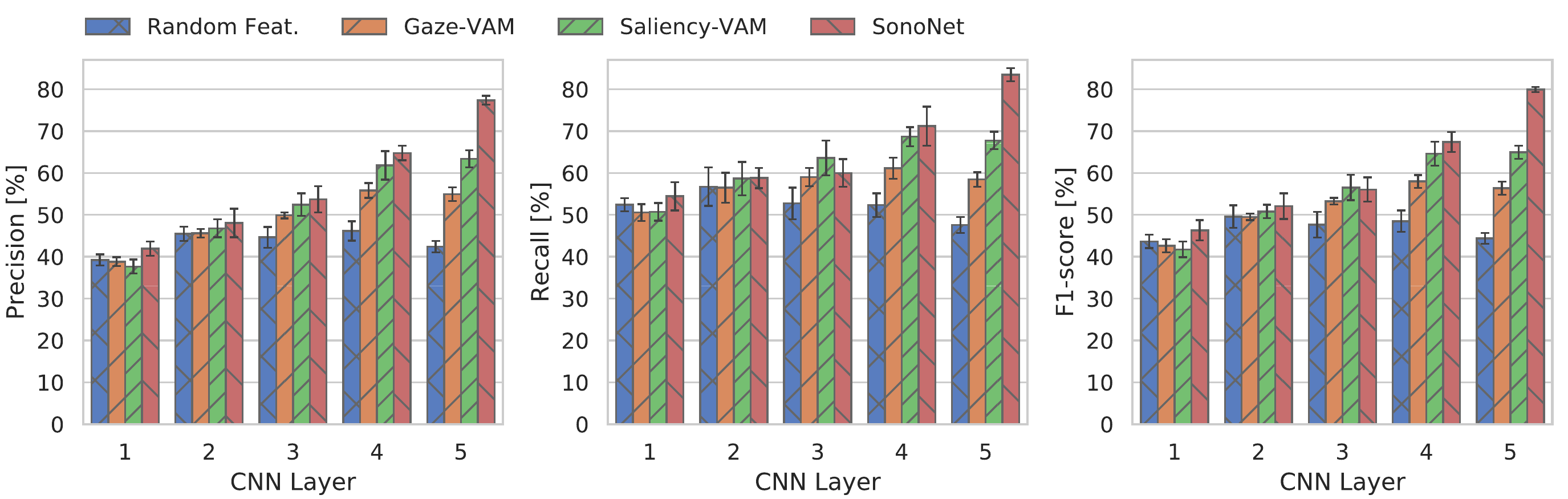}}
\put(0,101){a)}
\end{picture}\\[4mm]
\begin{picture}(347,108)
\put(0,0){\includesvg[width=\textwidth]{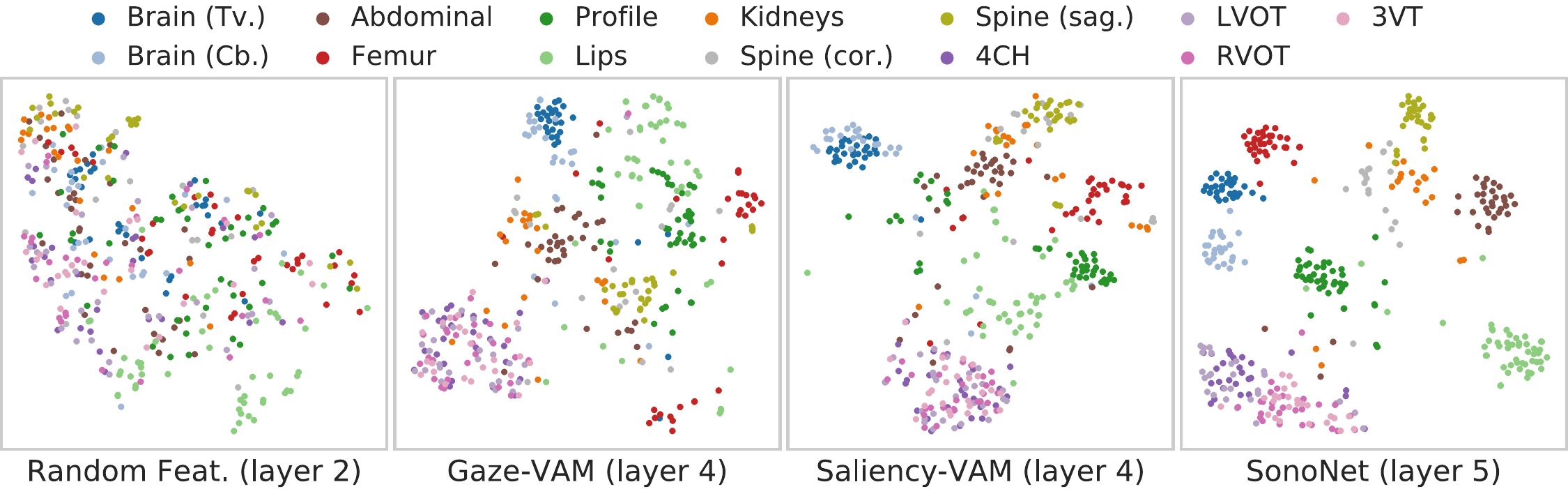}}
\put(0,100){b)}
\end{picture}
}
\caption{
a) Results of the regression analysis of the fixed-weight attention models, and baselines.
b) t-SNE visualization of the feature embeddings at the respective layers with the highest F1-score (Background class omitted for legibility).
Best viewed in color.
}
\label{fig:softmax}
\label{fig:tsne}
\end{figure}

\subsubsection*{Results.}
A quantitative evaluation of the fine-tuned attention models is shown in \autoref{tab:scores}.
The Saliency-FT model improves standard plane detection compared to the model trained from random initialization on every metric and for each standard plane.
The largest improvement per anatomy is observed for the right ventricular outflow tract (RVOT) with an average 20.8\% increase in F1-score, followed by the left ventricular outflow tract (LVOT) and the four chamber view (4CH).
Further, the average F1-score of Saliency-FT exceeds that of SonoNet on 3/14 classes.
The Gaze-FT model under-performs compared to training from random initialization.
In general, the average precision, recall and F1-score of SonoNet-FT are in good agreement with the literature values.
The authors do not provide per-anatomy scores for SonoNet-64.

The results of the regression analysis are shown in \autoref{fig:softmax} a).
The scores of both attention models monotonously increase up to layer 4 and stagnate at layer 5, peaking at F1-scores of $58.0\pm1.5\%$ for the Gaze-VAM and $64.9\pm1.5\%$ for the Saliency-VAM.
The scores of the Saliency-VAM and SonoNet are at similar levels up to layer 4, while the Gaze-VAM achieves lower scores.
For SonoNet the scores continue to increase at layer 5, reaching an F1-score of $79.9\pm0.6\%$.
In general, the scores of the random features are comparable to those of the other models at layers 0 and 1 but decline afterwards, peaking at an F1-score of $49.5\pm2.7\%$.

The differences between the feature embeddings are illustrated for selected layers in \autoref{fig:tsne} b) via t-SNE \cite{Maaten2008}, a non-linear dimensionality reduction algorithm that visualizes high-dimensional neighborhoods.
Compared to random features, a separation of the different standard plane classes emerges in the embeddings of the visual attention models.
However, a large overlap remains among the two brain views and the cardiac views, respectively.
Moreover, the views of coronal spine, kidneys, profile and lips are not well localized.
In the embedding of SonoNet, most classes are well-separated with overlap remaining among the cardiac views.

\section{Discussion and Conclusion}
The evaluations have shown that the visual attention models have learned meaningful representations of ultrasound image data, which was the main goal of this work.
In the transfer learning context, the Saliency-FT model clearly outperforms the model trained from random initialization.
The largest benefit is observed for the cardiac views with an average increase in F1-score of 15.3\%.
In fact, the performance of Saliency-FT is closer to that of SonoNet-FT, although the latter had been pre-trained with over 22k labeled standard plane images, while the attention models are pre-trained only with sonographer gaze on unlabeled video frames.
Since fine-tuning is performed with 753 standard plane images on average, this is a 30-fold reduction in the amount of manually annotated training data.
Gaze-FT did not yield an improvement, indicating that visual saliency prediction is better suited to learn transferable representations.

Even without fine-tuning, the high-level features of the attention models are predictive for fetal anomaly standard plane detection, outperforming the baseline with random weights for softmax regression on the feature activations.
Up to last network layer, the features of the Saliency-VAM are almost as predictive as those of SonoNet, even though it had received no explicit information about the concept of standard planes during training.
This confirms our hypothesis, motivated by Wu et al.~\cite{Wu2014}, that gaze is a strong prior for semantic information.
At the last layer, the attention models fall behind SonoNet, indicating the task-specificity of that layer.
The qualitative analysis through t-SNE confirms that some standard plane classes are well-separated in the respective feature spaces of the attention models, with overlap remaining for standard planes with similar appearance such as the brain views and the cardiac views, respectively.
It should be noted that we did not compare our models to a recently proposed variation of SonoNet \cite{Schlemper2018} with multi-layer attention-gating due to the added complexity of that architecture.

The results for both visual saliency prediction and gaze-point regression indicate successful learning of sonographer visual attention.
This is supported by the fact that the scores on the key metrics of AUC and NSS are higher than the scores reported by Cai et al.~\cite{Cai2018c} and than typical scores on the public MIT Saliency Benchmark of natural images (\url{saliency.mit.edu}).
However, our scores on the CC, SIM and KLD metrics are worse compared to these sources and in general, the comparability is very limited since the maximum attainable values are dataset-dependent.
For gaze-point regression, the proposed method based on the soft-argmax algorithm was found to be an effective solution.

In conclusion, we have shown that visual attention modeling is a promising method to learn image representations without manual supervision.
The trained CNNs generalize well to the task of fetal anomaly standard plane detection, both for transfer learning and as fixed feature extractors.
We have evaluated two methods for visual attention modeling, visual saliency prediction and gaze-point regression, and found that the representations learned with the former method generalize better.
The representation learning framework presented herein is generic and therefore has the potential to be applied in many settings where gaze and image data can be readily acquired.

\begin{small}
\textbf{Acknowledgements.}\quad
This work is supported by the ERC (ERC-ADG-2015 694581, project PULSE) and the EPSRC (EP/R013853/1 and EP/M013774/1).
AP is funded by the NIHR Oxford Biomedical Research Centre.
\end{small}

\end{document}